# Performance Analysis of Estimation of Distribution Algorithm and Genetic Algorithm in Zone Routing Protocol

Mst. Farhana Rahman
Computer Science and Engineering Discipline
Khulna University
Khulna 9208, Bangladesh
E-mail: farha_cseku@yahoo.com

S. M. Masud Karim
Computer Science and Engineering Discipline
Khulna University
Khulna 9208, Bangladesh
E-mail: masud@cse.ku.ac.bd

Kazi Shah Nawaz Ripon
Computer Science and Engineering Discipline
Khulna University
Khulna 9208, Bangladesh
E-mail: ripon@cse.ku.ac.bd

Md. Iqbal Hossain Suvo
Computer Science and Engineering Discipline
Khulna University
Khulna 9208, Bangladesh
E-mail: suvo_rgbd@yahoo.com

*Abstract*—In this paper, Estimation of Distribution Algorithm (EDA) is used for Zone Routing Protocol (ZRP) in Mobile Ad-hoc Network (MANET) instead of Genetic Algorithm (GA). It is an evolutionary approach, and used when the network size grows and the search space increases. When the destination is outside the zone, EDA is applied to find the route with minimum cost and time. The implementation of proposed method is compared with Genetic ZRP, i.e., GZRP and the result demonstrates better performance for the proposed method. Since the method provides a set of paths to the destination, it results in load balance to the network. As both EDA and GA use random search method to reach the optimal point, the searching cost reduced significantly, especially when the number of data is large.

*Keywords-Mobile Ad-hoc Network, Zone Routing Protocol, Estimation of Distribution Algorithm, Genetic Algorithm*

## I. INTRODUCTION

A *Mobile Ad-hoc Network* (MANET) is a collection of mobile nodes that dynamically form a temporary network. It forms the temporary network without any support of infrastructure. So, in the network there are possibilities of lack of reliability and unwanted delay. Again, if the number of nodes grows, the linear search will become costly and the complexity will become high. In case of large number of nodes, a random search will be beneficial where the worst case will equal the linear search. Because of frequently changing topology, low transmission power and asymmetric links routing protocols, MANET have to face the challenge for routing. *Zone Routing Protocol* (ZRP) is a widely used protocol for MANET. In 1997, ZRP was first introduced by Haas [2]. It was proposed to reduce the control overhead of proactive routing protocols and to decrease the latency caused by routing discover in reactive routing protocols.

Recently the scope of Genetic Algorithm (GA) has been extended to solve the ZRP problems. The GA has performed better in the sense of huge search space reduction, while guaranteeing the convergence of the solution. The GA is an adaptive heuristic search algorithm premised on the evolutionary ideas of natural selection and genetics [8]. The basic concept of GA is designed to simulate processes in natural system necessary for evolution. GA represents an intelligent exploitation of a random search within a defined search space to solve a problem. *Estimation of Distribution Algorithms* (EDA) [6], sometimes called Probabilistic Model-Building Genetic Algorithms (PMBGA), are an outgrowth of GA. In a GA, for an optimum solution a population of candidate solutions to a problem is maintained as part of the search. This population is typically represented as an array of objects. Here GA plays an important role in optimizing the search. This is because GA calculates the fitness of each population and generates a better population using crossover and mutation. So, the chance of getting good solutions increases dramatically. But due to the trap of local optima and the widespread diversity of solutions situations may occur where GA never converge to the optimal point that is failed to find a path which is existing between zones. And in some of the cases, GA takes longer time than expected to find a path.

This is the point where EDA works better than GA. Strictly speaking; GA and EDA are same apart from the crossover and mutation. There is nothing called crossover and mutation in EDA. Instead they use probabilistic model for generating new population. This guarantees the better generation of population than earlier generation. Also EDA converges faster, even if there is no feasible routing path. Thus the point is beneficial in the sense of performance of reduction in time and number of generations over GA by EDA. Here we prove the above; that is EDA finds routing path to a destination with minimum time and cost then GA from source when the source and destination is in different zone and number of node involved is large (about 100 to 1000 or more).

As in the mid to late 1990s, laptops and 802.11/Wi-Fi wireless networking became widespread, for research MANET became a popular subject. Many protocols have been proposed for routing in MANET. These protocols can broadly be classified into two types: *proactive* and *reactive* routing protocols. On case of proactive or *table-driven* protocol, by broadcasting routing updates in the network routes to all the nodes is maintain such as *Destination-Sequenced Distance Vector* (DSDV), whereas for reactive or *on-demand* protocols a route to the destination is determined only when the source attempt to send a packet to the destination such as *Dynamic Source Routing* (DSR). Using routing tables, proactive protocols maintain the routing information from one node to the other. Whenever the source has to send any packet to the destination, using the routing tables, path to destination can be found incurring minimum delay. But it may result in a lot of wastage of the network resources if a majority of these available routes are never used. Usually reactive protocols are associated with less control traffic. In a dynamic network a node has to wait until a route is discovered and a route discovery is expensive [7]. Also this causes unnecessary wastage of network resources and also wastage of time [5].

Hybrid protocols combine features of both reactive and proactive routing protocols. The ZRP is a hybrid protocol. It consists of proactive *Intra-zone Routing Protocol* (IARP), reactive *Inter-zone Routing Protocol* (IERP), and the *Border-cast Resolution Protocol* (BRP). ZRP works well both for table-driven protocols and on-demand protocols. But it provides short latency for finding new routes. Decision on the zone radius has significant impact on the performance. In ZRP, the actual problem comes when the destination is outside the zone. In this case, it makes use of Route Discovery with IERP, BRP and uses linear searching on the nodes. This process is time consuming and searching complexity arises as number of node involves increases [5] [7].

In order to detect new neighbor nodes and link failures, the ZRP relies on a Neighbor Discovery Protocol (NDP) provided by the Media Access Control (MAC) layer. NDP [9] transmits "HELLO" beacons at regular intervals. Upon receiving a beacon, the neighbor table is updated. Neighbors, for which no beacon has been received within a specified time, are removed from the table. If the MAC layer does not include a NDP, the functionality must be provided by IARP. Route updates are triggered by NDP, which notifies IARP when the neighbor table is updated. IERP uses the routing table of IARP to respond to route queries. IERP forwards queries with BRP. BRP uses the routing table of IARP to guide route queries away from the query source.

Recently GA has been used in MANET to find the optimized solution [1] [3] [4] [7]. A large amount of work has been done on the application of GA or evolutionary algorithms to communications networks.

Our objective is to use ZRP as an application in EDA, and compare the performance with the method used by the GA.

## II. LITERATURE REVIEW

### A. Zone Routing Protocol

The ZRP is based on the concept of zones [2]. For all the nodes in the zone, a routing zone is defined separately. The routing zone is based on the radius *r* which is then expressed in hops. Thus, the nodes included in the zone of a node are a maximum of radius *r* away from the node. In Fig. 1, the routing zone of *S* includes all the nodes from *A* to *I* but not *K*, as it resides further than the radius *r*. It should however be noted that the zone is defined in hops, not as a physical distance.

There are two types of nodes in a zone. The nodes residing with an exact distance of radius *r* are the peripheral nodes, and all the other nodes within the circles are interior nodes. The nodes are connected with each other bidirectional, if there is a routing path within the nodes. Intermediate nodes can be used to reach another node, based on the objective function. For example, in Fig. 1, we can reach node *H* from *S* by two possible ways; however only one route is chosen based on the objective criterion. A detail of ZRP can be found in [2] for further reading.

### B. Genetic Algorithm

GA [8] is an evolutionary approach to reach to an optimal point in a search space. For larger search space, GA becomes more meaningful and it reduces the searching time, explores in various dimensions within the search space using different GA techniques, like *crossover*, *mutation* etc. Although there are possibilities to trap in the local optima, there are several ways of getting out of it using crossover and thus reach global optima.

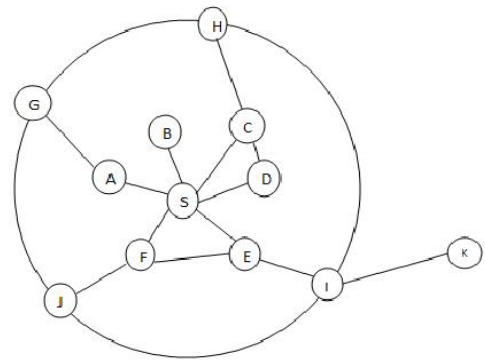

Figure 1. Routing zone of *S* with *r* = 2.

The outline of basic GA with description is given below:

1. **[Start]** Generate random population of $n$ chromosomes (suitable solutions for the problem)
2. **[Fitness]** Evaluate the fitness $f(x)$ of each chromosome $x$ in the population
3. **[New population]** Create a new population by repeating following steps until the new population is complete
   (a) **[Selection]** Select two parent chromosomes from a population according to their fitness (the better fitness, the bigger chance to be selected).
   (b) **[Crossover]** With a crossover probability cross over the parents to form new offspring (children). If no crossover was performed, offspring is the exact copy of parents.
   (c) **[Mutation]** With a mutation probability mutate new offspring at each locus (position in chromosome).
   (d) **[Accepting]** Place new offspring in the new population.
4. **[Replace]** Use new generated population for a further run of the algorithm.
5. **[Test]** If the end condition is satisfied, **stop**, and return the best solution in current population.
6. **[Loop]** Go to step **2**.

As we can see from the GA outline, the *crossover* and *mutation* are the most important parts of the algorithm. The performance is influenced mainly by these two operators. Detail of GAs can be found in [8].

*C. Estimation of Distribution Algorithm*

In EDAs [6], the problem specific interactions among the variables of individuals are taken into consideration. It is the most recent adaptation of evolutionary approaches. It is starting to be widely used as a promising alternative of GA. The evolving process of EDA is the same as GA apart from crossover and mutation. Instead, EDA uses probabilistic distribution. The probability distribution is calculated from a database of selected individuals of previous generation. The pseudo code of EDA can be formulated as follows:

1. **[Start]** $D_0 \leftarrow$ Generate $M$ individuals (the initial population) at random.
2. **[Fitness]** Evaluate the fitness $f(x)$ of each chromosome $x$ in the population. Repeat steps 3 to 5 for $l = 1, 2 \ldots$ until the stopping criteria met.
3. **[Selection]** $D_{l-1}^{se} \leftarrow$ Select $N <= M$ individuals from $D_{l-1}$ according to selection method.
4. **[Estimation]** $p_l(X) = p(X|D_{l-1}^{se}) \leftarrow$ Estimate probability distribution of an individual being among the selected individuals.
5. **[New Population]** $D_l \leftarrow$ Sample $M$ individuals (the new population) from $p_l(x)$.

The easiest way to calculate the estimation of probability distribution is to consider all the variables in a problem as univariate. Then the joint probability distribution becomes the product of the marginal probabilities of $n$ variables, i.e.,

$$p_l(x) = \prod_{i=1}^{n} p(x_i). \tag{1}$$

*D. Univariate Marginal Distribution Algorithms*

In UMDA [6], it is assumed that is there is no interrelation among the variables of the problems. Hence the $n$-dimensional joint probability distribution is factorized as a product of $n$ univariate and independent probability distribution. That is:

$$p_l(X) = p(X|D_{l-1}^{se}) = \prod_{i=1}^{n} p(x_i). \tag{2}$$

The pseudo code for UMDA is as follows:

1. $D_0 \leftarrow$ Generate $M$ individuals (the initial population) at random.
2. Repeat steps 3 to 5 for $l = 1, 2\ldots$ until stopping criteria met.
3. $D_{l-1}^{se} \leftarrow$ Select $N \leq M$ individuals from $D_{l-1}$ according to selection method.
4. Estimate the joint probability distribution

$$p_l(X) = p(X|D_{l-1}^{se}) = \prod_{i=1}^{N} p(x_i). \tag{3}$$

5. $D_l \leftarrow$ Sample $M$ individuals (the new population) from $p_l(x)$.

In UMDA the joint probability distribution is factorized as a product of independent univariate marginal distribution, which is estimated from marginal frequencies:

$$p_l(x_i) = \frac{\sum_{j=1}^{N} \delta_j(X_i = x_i | D_{l-1}^{se})}{N} \tag{4}$$

with $\delta_j(X_i = x_i | D_{l-1}^{se}) = 1$, if in the $j$th case of $D_{l-1}^{se}$, $X_i = x_i$; 0 otherwise.

III. PROPOSED METHOD

We choose a random network with maximum chromosome length $N$, and applied ZRP to determine different zones with a radius of $r$, where $r$ is the maximum distance of a node from the central node of a zone. This gives us simplified form of a route from the source node to the destination node using border nodes. Using this route as a chromosome, we create the population and apply GA and EDA. The GZRP uses the popular encoding scheme and the minimizing fitness function.

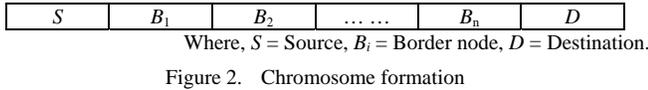

Where, $S$ = Source, $B_i$ = Border node, $D$ = Destination.

Figure 2. Chromosome formation

Here, all the nodes belong to different zones. The initial population is created randomly, containing the individuals of the above chromosome format. The minimizing fitness function would be the one with finding the shortest route from the source to the destination. The function can be given as follows:

$$I_{ij} = \begin{cases} 1 & \text{if the link from node } i \text{ to node } j \text{ exists} \\ 0 & \text{otherwise} \end{cases} \quad (5)$$

Thus, we choose the objective function as the cost of two interconnected nodes multiplied by $I_{ij}$. In case of GA, we apply one-point crossover and mutation to get rid of stacking in local optima and increase the diversity of solutions. The point is chosen randomly, and crossover is applied between the two randomly selected individuals. Mutation operator then flips the randomly selected genes of the newly formed chromosome with the partial route from the mutation point.

In case of EDA, we use the same encoding scheme and the fitness function. As there is no crossover and mutation in EDA, the only challenge was to compute the probabilistic distribution of chromosome. The problem in this case is trivial. All the chromosome lengths are not the same, from the source node to the destination nodes. So, we apply the technique of continuous EDA domain, where the probabilistic model is generated using the mean and standard deviation. Thus the random function used in this case is the normal distribution function. In both the experiments, we use two terminating criterion, namely, maximum number of round in a single run and the converged solutions. Whenever we reach a converged solution, the program terminates, and if the program cannot converge to an optimal solution, we stop the run after a fixed number of iteration.

## IV. EXPERIMENT AND RESULT ANALYSIS

In our proposed method, we use the maximum number of iteration in a single run as 1000. The network length varies from 100 to 1000. The sub-population size used in EDA is 50 percent of the main population. The mutation factor is used as 90 percent, meaning a high probability of mutation chance for each individual. As we want to apply our method to ZRP, we do not use any benchmark data set of networks; rather try to handle the situation of dynamically formed network. Thus we increase the network size from 100 to 1000 with the increment of 100 nodes each time. Then we run the program for each set 10 times and used the average of the solution.

Figure 3 shows the performance of GA and EDA in terms of required number of Generations. This figure gives a clear view that, for the above parameter settings, EDA outperforms GA when the network grows in size. For the simplicity in implementation, we considered only the cost of the routing path in a zone.

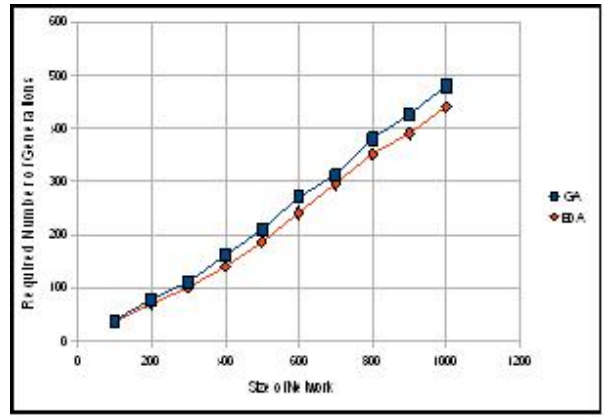

Figure 3. Required number of generation to find the converged value for the same network size using GA and EDA. *(figure caption)*

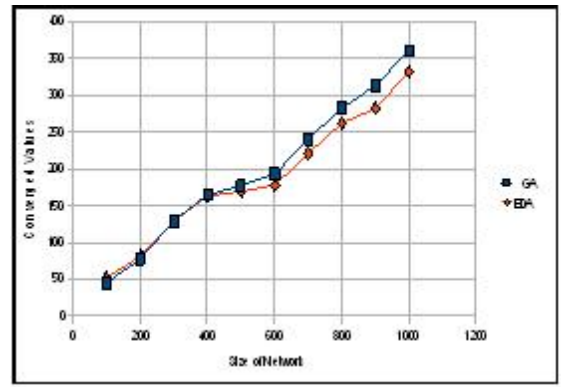

Figure 4. Converged values determined by GA and EDA.

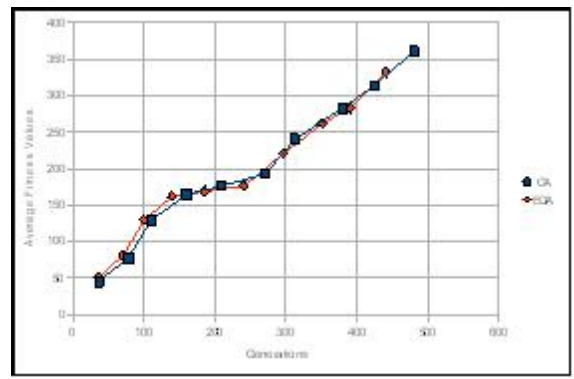

Figure 5. Average fitness values determined by GA and EDA of 50 individual runs.

In Figure 4 the best value of the 50 individual runs is taken to measure the performance against the average number of generations. Here it can be seen that GA performs better when the network size is small. In this case, when the network size grows more than 400 EDA performs better by resulting in a lower converged value for large size network. Thus our approach of applying EDA to solve ZRP proved to perform better than GA. Figure 5 states the average fitness values (in this case the optimal routing path cost) in each generation by EDA and GA.

Again, as our objective function was minimizing, the lower values of EDA indicates better solutions over GA. Also we are obtaining a set of routing path from source to destination from generation. As the network is generated randomly the same routing path is not used as the shortest path from source to destination which results in low traffic and load balance in the network.

## V. CONCLUSION

Evolutionary approaches are not guaranteed to find the optimal solutions but they can minimize the cost significantly and are proved effective in larger search space. EDA is a growing field in evolutionary approaches and becoming popular day by day. Our contribution opens a new scope of applying EDA in such a field like ZRP, where GA is already applied. In this study, we only consider UMDA. In future we can extend our work to apply population based incremental learning (PBIL) algorithm [6] and Compact Genetic Algorithm (CGA) [6] both of which are forms of EDA. Then we can decide the best EDA approach to solve ZRP. Again, the path rediscovery can be solved in case of a break down in the network by EDA and GA. Thus, we can again compare the performance in this aspect.

AUTHORS PROFILE

Mst. Farhana Rahman was an undergraduate student of Computer Science and Engineering (CSE) Discipline, Khulna University, Bangladesh. She has started his B.Sc.Engg.(CSE) degree in 2005. She did his undergraduate thesis in the field of evolutionary computing. She has particularly shown his keen interest in zone routing protocol in Mobile Ad-hoc Network.

S. M. Masud Karim has been serving as a faculty member of Computer Science and Engineering (CSE) Discipline, Khulna University, Khulna, Bangladesh. He completed his B.Sc.Engg.(CSE) degree with distinction in 2001. He went abraod for hisher studies in 2006 and was awarded M.Sc. in Media Informatics from Technical University of Aachen (RWTH Aachen), Germany in 2008 and M.Sc. in Informatics from University of Edinburgh, UK in 2009. His areas of interest include information retrieval, data exchange, data integration, computer security.

Kazi Shah Nawaz Ripon has been serving as a faculty member of Computer Science and Engineering (CSE) Discipline, Khulna University, Khulna, Bangladesh. He completed his B.Sc.Engg.(CSE) degree with distinction in 2000. He completed M.Phil in Computer Science from the City University of Hong Kong in 2006. He is currently doing his Ph.D in the University of Oslo, Norway. His areas of interest include multiobject evolutionary alogirthms, genetic algorithm and computer network.

Md. Iqbal Hossain Suvo is an undergraduate student of Computer Science and Engineering (CSE) Discipline, Khulna University, Bangladesh. He has started his B.Sc.Engg.(CSE) degree in 2005. He did his undergraduate thesis in the field of evolutionary computing.